\documentclass{article}

\usepackage{arxiv}

\usepackage[utf8]{inputenc} 
\usepackage[T1]{fontenc}    
\usepackage{hyperref}       
\usepackage{url}            
\usepackage{booktabs}       
\usepackage{amsfonts}       
\usepackage{nicefrac}       
\usepackage{microtype}      
\usepackage{lipsum}
\usepackage{graphicx}
\usepackage{array}
\usepackage{caption}
\usepackage{lscape}
\usepackage{listings}
\usepackage{xcolor}
\graphicspath{ {./images/} }

\title{Meta Arcade: A Configurable Environment Suite for Meta-Learning}
\author{

Edward W. Staley, Chace Ashcraft, Benjamin Stoler, Jared Markowitz, 
\\ \textbf{Gautam Vallabha, Christopher Ratto, Kapil D. Katyal} 
\\
\\
Johns Hopkins University Applied Physics Laboratory
}

\begin{document}
\maketitle
\begin{abstract}
Most approaches to deep reinforcement learning (DRL) attempt to solve a single task at a time. As a result, most existing research benchmarks consist of individual games or suites of games that have common interfaces but little overlap in their perceptual features, objectives, or reward structures. To facilitate research into knowledge transfer among trained agents (e.g. via multi-task and meta-learning), more environment suites that provide configurable tasks with enough commonality to be studied collectively are needed. In this paper we present Meta Arcade, a tool to easily define and configure custom 2D arcade games that share common visuals, state spaces, action spaces, game components, and scoring mechanisms. Meta Arcade differs from prior environments in that both task commonality and configurability are prioritized: entire sets of games can be constructed from common elements, and these elements are adjustable through exposed parameters. We include a suite of 24 predefined games that collectively illustrate the possibilities of this framework and discuss how these games can be configured for research applications. We provide several experiments that illustrate how Meta Arcade could be used, including single-task benchmarks of predefined games, sample curriculum-based approaches that change game parameters over a set schedule, and an exploration of transfer learning between games.
\end{abstract}

\section*{Acknowledgements}
Primary development of Meta Arcade was funded by the DARPA Lifelong Learning Machines (L2M) Program. Additionally, this work relates to Department of Navy award N00014-20-1-2239 issued by the Office of Naval Research. The United States Government has a royalty-free license  throughout the world in all copyrightable material contained herein.

\section{Introduction}

Early deep reinforcement learning (DRL)~\cite{dqn2013, trpo, ddpg, alphazero, ppo} algorithms focused on performance on very narrowly-defined tasks.  Only more recently have researchers looked to extend learned behaviors outside of the training domain or to learn several tasks simultaneously. Accordingly, existing benchmark environments for DRL tend to be narrow in scope. For example, DQN \cite{dqn2013}, one of the first works to combine deep neural networks and reinforcement learning, played individual Atari 2600 games using the Atari Learning Environment (ALE) \cite{bellemare:arcade}. While these games share the Atari name, they have little else in common, making them unsuitable for multitask learning or meta-learning (as noted in \cite{metaworld}).

There is a considerable gulf between learning individual games and truly general learning. Humans are capable of learning common skills across tasks, improving their abilities in a general sense, and quickly adapting to novel situations. A variety of research domains, including transfer learning, multitask learning, meta-learning, and lifelong learning, have sought to expand the capability of DRL algorithms beyond a single task. However, there is a need for evaluation environments to support this work. Moving away from a narrow training domain remains a very difficult challenge, requiring environments that allow a graceful transition to multiple tasks and expose precise control over the underlying distributions.

In this work we present Meta Arcade, an attempt to address this need. Meta Arcade is a suite of lightweight environments designed for multi-task learning and meta-learning, with an emphasis on configurability. Game elements, colors, dynamics, and objectives are all parameterized with support for distribution sampling, such that a single game can be varied slightly or altered to the extent that it constitutes a new task. For example, consider a task similar to Atari's Breakout, in which the paddle, ball, and blocks have sizes, speeds, and colors that can all be defined as constants or distributions, allowing for a nearly infinite spectrum of Breakout tasks. Common game objectives like collecting blocks, bouncing balls, and avoiding obstacles are re-used between games. If multiple tasks are defined, these common objectives will result in clear overlap between the games, making them viable for multitask approaches. 

Our main contributions in this work are the Meta Arcade framework and suite, which provide two major assets to researchers.  First, they provide an environment for RL that is fully parameterized so that it can be tuned to the needs of individual studies, including examinations of task variation or domain shift as a direct focus. Second, Meta Arcade provides this capability for not only a single task, but for an entire space of arcade-inspired tasks that have common parameters, enabling research that requires sets of distinct but related tasks. We have defined 24 unique games that can be constructed using Meta Arcade, and describe how others may be built. Meta Arcade will be available on github (release in progress) with accompanying documentation and examples.  Additionally, we contribute a series of experiments that include acquiring single-task expertise, curriculum-based learning, and multitask learning.

This manuscript details the capabilities of Meta Arcade and the results of our experiments. In Section \ref{related} we discuss other DRL environments and how they support the current research landscape. In Section \ref{metaarcade} we describe Meta Arcade in detail and how it can be used. Benchmarks and other experimental results are presented in Section \ref{experiments}, and a discussion of future work is presented in Section \ref{future}.

\section{Related Work} \label{related}
There are a variety of environments that currently support DRL research, both for single tasks and, to a lesser extent, for multiple tasks. In addition to the discrete setting of ALE, environments for continuous-space control problems have been created. DDPG~\cite{ddpg} introduced several 2D control tasks and used MuJoCo \cite{mujoco} as a platform for building such environments. Later MuJoCo was used to build 3D environments, as introduced alongside generalized advantage estimation \cite{gae}. ALE, MuJoCo environments, classic control environments, and others are maintained in the OpenAI Gym \cite{gym}.

While these benchmark environments are critical to testing and evaluating single-task experts, they tend to present narrow tasks, which often lead to policies that are hyper-fit to a specific problem \cite{overfitting, actormimic}. This concern can be addressed algorithmically, but may be more naturally addressed by broadening the scope of the environments.  One common avenue towards robustness is to envision a task not as a static problem but rather as a distribution around task parameters that may better encompass the inherent randomness of reality, or enable sim-to-real transfer by presenting a super-set of features that may be encountered. The environment can be randomized in any number of ways to encourage more robust learning, from visual distortions to changing dynamics \cite{simtoreal}.  This general idea, termed domain randomization, was first presented for supervised learning in \cite{domain_rand} and laid the groundwork for policy-based success in the robotic manipulation of a Rubik's Cube \cite{rubiks}. Similar ideas of a learning a single task defined by a large distribution have been accomplished entirely in the physical world \cite{armfarm} or built into simulation via procedural generation. Notably, Procgen \cite{procgen}, which includes the game Coin Run \cite{coinrun}, defines tasks that are built procedurally and have key visuals randomized.

If a target environment is difficult to learn but can be modified to create more easily-learned variants, it can be useful to supply a learning curriculum~\cite{curriculum,  MACALPINE201821, narvekar2020curriculum, robot_curriculum}. Often a curriculum occurs naturally as a part of training, especially in self-play (e.g. AlphaZero \cite{alphazero}), but this is rarely the case in single-player games.  If the adversary is the environment itself, then creating a curriculum requires a configurable environment. Several environments provide options towards customization or even tend towards being a framework for building scenarios. ViZDoom \cite{vizdoom} and DeepMind Lab \cite{dmlab} both provide tools for creating custom challenges that maintain the fundamental mechanisms of the games, which could be used to create a series of challenges with overlapping elements. However, it is difficult to build out new tasks for these platforms, and ultimately all scenarios look more like variations of the original game than distinct tasks in their own right. Explicit simulation frameworks like Unity \cite{mlagents} and MuJoCo provide the means to create entirely new environments or modify existing examples, but require full development efforts. 

Multi-task learning examines training competent agents on several tasks simultaneously or sequentially, rather than solely on instances of a single task. Ideally, the different tasks have features or skills that overlap such that there is an advantage to learning them together~\cite{actormimic, metaworld}.  An open research question is how to measure such similarity so that an algorithm can organize or prioritize the tasks in some way~\cite{task_sim4libraries2005, sparse_coding2012, ammar2014automated, zhu2020transfer}. While a suite of environments is needed for multi-task RL research, ensuring that tasks are sufficiently similar or have sufficiently overlapping skills to enable transfer is difficult. Works such as Meta-World \cite{metaworld} and the Sonic benchmark \cite{sonic} have been designed with this requirement in mind. Suites which are based around a common game, such as the StarCraft II Learning Environment \cite{sc2} minigames, tend to inherently include this task overlap. This overlap is also useful if the tasks will be learned in sequence, as in lifelong learning and its variations. Alchemy \cite{alchemy} takes these ideas further by re-sampling the underlying game structure each episode, such that every episode is a different draw from a common task distribution.

Finally, another approach to producing more generalizable agents is meta-learning, or ``learning-to-learn''~\cite{schmidhuber1987learningtolearn, learningtolearn_survey}, which enables adaptation to new tasks in a structured way.  Meta-learning has considerable overlap with the ideas discussed above but is more explicitly interested in this adaptation: the learning objective may be to determine what knowledge is transferable between tasks, to learn how to transfer that knowledge, or to learn how to do so efficiently, with the goal of quickly becoming proficient at a new task. Current research questions in this area include how to apply common meta-learning techniques \cite{maml, reptile, pearl} to problems with discrete action spaces and to problems with sparse rewards. Meta-World \cite{metaworld} and Alchemy \cite{alchemy} are two recent environments that were specifically designed to support meta-learning research, and interestingly both mention that while the popular benchmarks of arcade games (as in \cite{bellemare:arcade}) provide a large array of tasks with diverse structure, this diversity is so wide as to be an impediment to the shared learning of multiple tasks. The goal of this work is to enable further exploration of multitask and meta-learning by providing game environments that are useful for these burgeoning research areas. Meta Arcade has the variety, speed, and ease of use that accompanies arcade games as a research benchmark, while providing the rich configurability and shared task structure needed for multitask learning, meta-learning, and related topics.

\section{Meta Arcade} \label{metaarcade}

\subsection{Environment Suite}

The Meta Arcade environment suite seeks to facilitate many of the above research directions through a highly flexible set of tools for training and evaluating sophisticated DRL algorithms.  A primary focus was to create a suite of arcade games with many areas of overlap, such that the collection has a diversity of tasks similar to ALE while having enough commonality among tasks to be suitable for multi-task learning and meta-learning. Each game is parameterized and can be easily modified or extended, as is discussed in the next section.

\begin{figure}[ht!]
\centering
  \includegraphics[width=\linewidth]{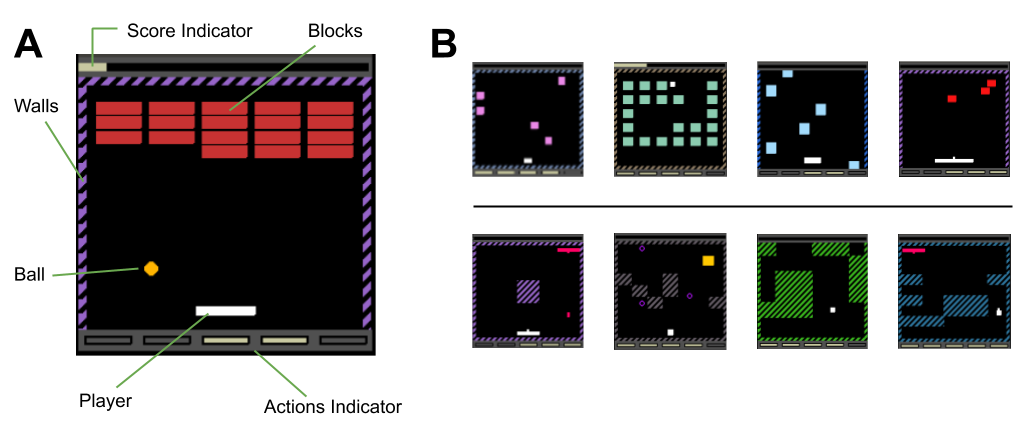}
  \captionsetup{width=0.95\linewidth}
  \caption{\small \emph{Some features of Meta Arcade.} \textbf{A:} The game \emph{breakout} with some highlighted elements, including the visual indicators that are outside of the game play area. \textbf{B:} Examples of common elements among games. The top row shows some games that feature solid-color collectable blocks, while the bottom row shows static walls that have a common visual texture.  Several other elements are seen throughout the game suite, such as hazards, balls, or opponents.}
  \label{fig:game_features}
\end{figure}

Meta Arcade includes 24 predefined games that have the following characteristics: All games present a 84x84x3 pixel-based state space, and all games have identical action spaces (6 discrete actions). Some games will not use all 6 actions, but will instead map unused actions to no-ops internally such that the full action space is preserved.  The actions correspond to [No-op, Up, Down, Left, Right, Shoot] and have identical effects in all games where they are used. Continuous actions are also supported as an option. A similar entity is controlled by the player in all games. An overview of the game display is shown in Figure \ref{fig:game_features}. Meta Arcade was built with Pygame\footnote{\url{https://www.pygame.org}}, a platform-agnostic Python API for video game development, making Meta Arcade immediately accessible to developers already using Python deep-learning frameworks. Meta Arcade can optionally be run headlessly.  

Games are constructed from a common set of game mechanics with common visuals and scoring systems. In general, if a game element appears in more than one game, it will have similar interactions with other objects and incur similar rewards (if applicable).  Games are designed to have a variety of overlapping skills such that a policy to solve one game is potentially useful for others. For example, multiple games involve a paddle and a ball while others involve navigation in four directions.  A sampling of games with some highlighted commonalities can be seen in Figure \ref{fig:game_features}.

The maximum score for all games is 100 points and the minimum score is either 0 or -100, depending on whether it is possible to lose the game (as opposed to the game timing out, which happens after several thousand frames). Rewards are event-driven and sparse. Common scoring mechanisms are built around this scale and our summarized in the table below. More than one scoring mechanism can be present in a game.

\begin{center}
\begin{tabular}{ |c|c| } 
    \hline
    Game Event & Score \\
    \hline
    \hline
    Player passes opposite side & +100 (win) \\
    \hline
    Player collides with collectable block & + Block Value \\
    \hline
    Player collides with hazard & -100 (loss) \\
    \hline
    Ball passes opponent side & +100 (win) \\ 
    \hline
    Ball passes player side & -100 (loss) \\
    \hline
    Ball collides with collectable block & + Block Value \\
    \hline
    Bullet collides with collectable block & + Block Value \\
    \hline
    Bullet collides with opponent & +100 (win) \\
    \hline
    Bullet collides with player & -100 (loss) \\
    \hline
    Block falls passed player side & -100 (loss)\\
    \hline
\end{tabular}
\end{center}
\captionsetup{width=0.95\linewidth,skip=0.2cm}
\captionof{table}{Reward mechanisms in Predefined Meta Arcade Games}
\bigskip

There are no lives in Meta Arcade, as past approaches to solving arcade games often manually split episodes among lives anyway \cite{dqn2013, dqn2015}. Block-collecting games divide 100 points among some target number of blocks. This leads to some initially unintuitive results for single games in which a loss is heavily weighted against progress.  For example, collecting 19 out of 20 blocks in breakout results in a score of -5 (+5 points x 19 and -100 for missing the ball).  This does not prevent learning of the games but does require care when examining results. For example, converging to a score of 0 indicates successful learning for some games, and high variance in the final performance may be normal. We believe this is a reasonable tradeoff in exchange for a common scoring system among all games, which may aid in the learning of value functions with similar magnitudes.

The predefined games in Meta Arcade present a range of difficulties; some are quickly solvable with off-the-shelf DRL algorithms while others are challenging for even customized approaches. This spectrum is presented intentionally to provide easy games for rapid testing and challenge games to motivate future research. A few games are quite rudimentary and are meant to demonstrate a specific mechanic with readily available rewards.  Others that are more challenging may use more than one skill or mechanism, provide sparse reward signals, or both. Not all games are necessarily solvable in isolation, and may require curricula, transfer learning, or other advanced methods to solve.

\subsection{Capabilities for Modern DRL Research}

While many environments are presented to the research community as static problems, there is often a close collaboration in the creation of new algorithms and new environments. Environments may need to be specially built to explore a research idea, or a novel environment may spark a new research direction. With this in mind, Meta Arcade games are editable and are designed to be tweaked as needed for a given research avenue. While many predefined games are supplied, Meta Arcade can be thought of as a game creation tool as much as a suite of specific challenges.  New games can easily be created from scratch or by editing an existing game.

Each game is entirely defined by a JSON file (see Figure~\ref{fig:json_game} in the appendix) which lists the relevant game mechanics and their attributes. The sizes, speeds, colors, and behaviors of any game can be directly changed through these files, or modified using environment tools from python scripts.

\begin{figure}[ht!]
\centering
  \includegraphics[width=\linewidth]{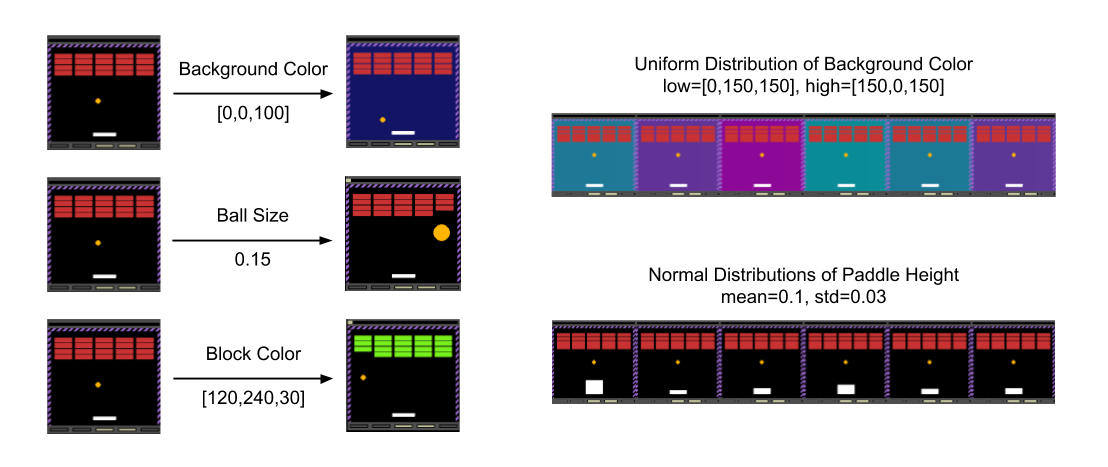}
  \captionsetup{width=0.95\linewidth}
  \caption{\small \emph{Parameter changes to} breakout. All colors, sizes, speeds, and other behaviors can be easily modified to customize games for experimentation. Some static value changes are shown on the left, some of which are purely visual while one changes the gameplay itself. Almost all parameters can be replaced with distributions which are re-sampled each episode. The effects of using distributions can be seen in the examples on the right.}
  \label{fig:game_params}
\end{figure}

Furthermore, any value in the game definitions can be replaced with a distribution to be sampled from rather than a static value. Gaussian distributions, uniform distributions, and special color distributions are supported. This leads to exploration of task distributions, domain randomization, and domain shifts. Some examples of parameter changes can be found in Figure \ref{fig:game_params}. New games can be created by simply creating a new JSON parameter file.

\begin{figure}[ht!]
\centering
  \includegraphics[width=\linewidth]{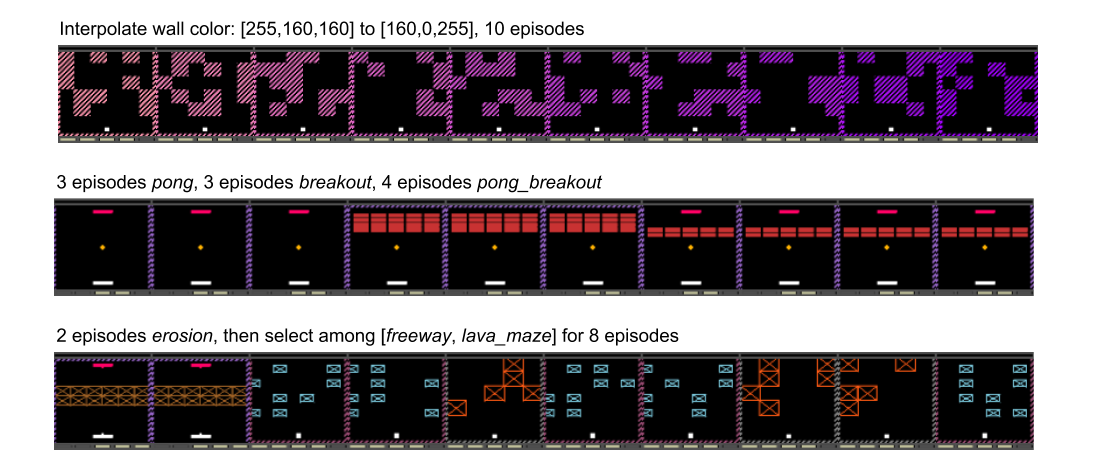}
  \captionsetup{width=0.95\linewidth}
  \caption{\small \emph{Simple example curricula using built-in tools.} A variety of curriculum features are available to create training regimens from Meta Arcade. The examples above are limited to 10 episodes for demonstration purposes, but long sequences of tasks and their variations can be constructed. Curricula can optionally be defined in terms of steps rather than episodes.}
  \label{fig:simple_curricula}
\end{figure}

Additionally, Meta Arcade includes some simple tools for building training curricula based on multiple tasks, or many copies of a task with shifted parameters. Games may be grouped together in a pool of tasks to be sampled, repeated for a specific number of episodes or steps, or interpolated between in the configuration parameter space. Together with access to game parameters, this enables the construction of tasks which may change over time, may be interwoven with other tasks, or may be combined in other manners. This type of functionality is essential to explore areas like continual learning, transfer learning, and curriculum learning.  Some examples of the environment’s behavior under simple curricula can be seen in Figure \ref{fig:simple_curricula}.

The most powerful and nuanced feature of the curriculum tools is the ability to linearly interpolate between game parameters, including those defined by distributions. Over a set number of episodes, two games defined by identical JSON fields can be interpolated between at each reset call. If a scalar value is supplied, it is interpolated over time.  Integer values are interpolated as floating-point and then rounded back to the nearest integer. If a distribution is specified, the parameters of that distribution are interpolated. For example, the standard deviation of an object’s size could be changed over time, or the number of blocks could be sampled from a uniform distribution that has changing bounds. This is incredibly powerful for slowly changing the difficulty of a game or controlling domain randomization as a function of time.

Finally, some image post-processing is included in the JSON parameter files and may be changed as a fundamental aspect of the game. The hue, saturation, and value of the image can be adjusted. There is also an option for color inversion and image rotation. These tools are useful to quickly produce multiple variants of a game for studying a distribution of visual representations on a common task.  For example, rather than changing the color of several game elements, a simple hue shift can affect the entire image.  As with all other parameters, these image alterations can be replaced with distributions and interpolated between. 

\section{Experiments and Results} \label{experiments}
To demonstrate the range of features available in Meta Arcade and their applications to DRL research, we conducted a series of experiments which benchmark and test different games and RL approaches. These experiments can be broken into three sections: (1) Single-task benchmarking of all predefined games using PPO \cite{ppo} (This also serves as a summary of all available games), (2) Selected curricula-based approaches for challenging games and the exploration of domain distributions, and (3) Transfer learning experiments on subsets of the games explore utilizing their common features. These three sections are described in more detail below. We have chosen to discuss the results after each subsection since the results of each experiment often influenced the design of the next.

\subsection{PPO Benchmarking}

\subsubsection{Benchmarking Experiments}
To demonstrate the solvability of predefined games and provide a point of reference for other researchers, we used our own implementation of Proximal Policy Optimization (PPO) \cite{ppo} to try to learn each game separately. This is useful not only for understanding the difficulty of a given game but also proved useful for designing the games themselves and even debugging.  In an effort to provide a range of reasonable difficulty, several games were tweaked in an iterative process with PPO benchmarking. In some respects, this experiment could be considered part of the environment design process.

We trained PPO once on each game for 10 million frames, using 8 workers distributed with the Message Passing Interface (MPI) \cite{mpi, mpi4py}. Gradients are computed in each worker and averaged in the main process.  A standard network architecture for Atari was used \cite{dqn2013}. Specific hyperparameters can be found in the appendix, Section \ref{ppoparams}.

\subsubsection{Benchmarking Results}
The results from benchmarking all 24 games can be clearly split into those that were successful and unsuccessful, with "unsuccessful" meaning that trained agent failed to consistently reach a positive score. 18 of the 24 were successfully learned, often with nearly maximal performance. The training results of the successful games are summarized in Table \ref{table:1}. The only games which were not clearly successful in this category are Juggling, Keep Ups, and Pong Breakout.  Due to the severe penalty of missing the ball, Juggling and Keep Ups are actually very performant and so the results curve is somewhat deceptive. They have not reached a maximum average score but are still considered successful. Similarly, Pong Breakout is imperfect but still doing well; a score of 71 in this zero-sum case converts to a win rate of about 85\% against the opponent.

The six unsuccessfully learned games can be viewed in Table \ref{table:2}.  These games tended to converge to average scores around -100 or 0, corresponding to complete failure or avoiding play entirely.  These games appear to be challenging because it is very hard to discover rewards naturally or without incurring some penalty, which ultimately discourages learning the needed behaviors for success. This is a common problem in reinforcement learning with sparse rewards or competing reward signals. In the next section, we describe how Meta Arcade’s parameterization and curriculum-building tools can be leveraged to explore these areas.

\newcommand{\tpicwidthA}{1.8cm}
\newcommand{\tpicwidthB}{2.8cm}

\begin{table}[h!]
\centering
\begin{tabular}{|>{\footnotesize\centering\arraybackslash}m{2.4cm}|>{\footnotesize\centering\arraybackslash}m{3cm}|>{\footnotesize\centering\arraybackslash}m{1cm}||>{\footnotesize\centering\arraybackslash}m{2.4cm}|>{\footnotesize\centering\arraybackslash}m{3cm}|>{\footnotesize\centering\arraybackslash}m{1cm}|} 
 \hline
 Game & Training Curve & Score & Game & Training Curve & Score\\
 \hline\hline
 
 Avalanche \includegraphics[width=\tpicwidthA]{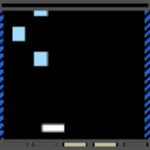} & \includegraphics[width=\tpicwidthB]{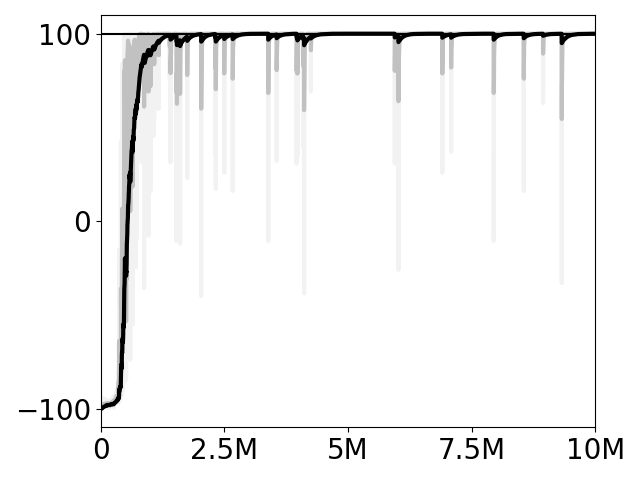} & 100.00 $\pm$ 0.00 & 
Juggling \includegraphics[width=\tpicwidthA]{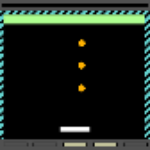} & \includegraphics[width=\tpicwidthB]{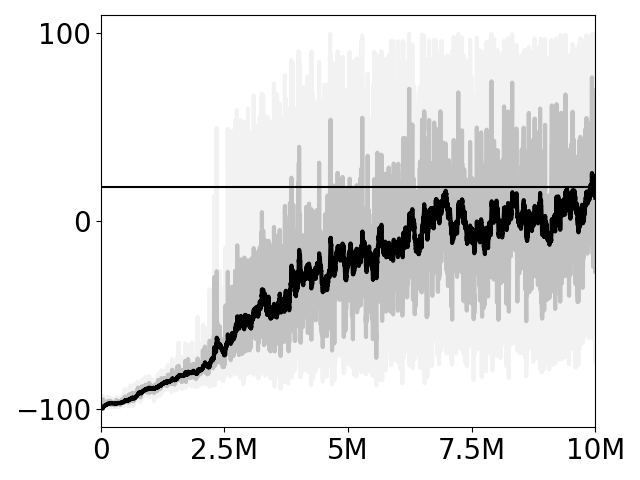} & 17.98 $\pm$ 68.33 \\ 
 \hline

 Battle Pong \includegraphics[width=\tpicwidthA]{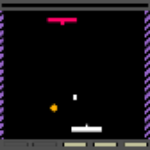} & \includegraphics[width=\tpicwidthB]{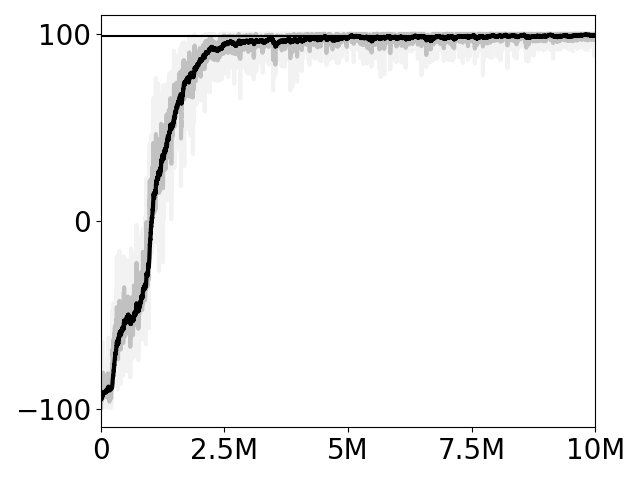} & 99.03 $\pm$ 3.00 & 
 Keep Ups \includegraphics[width=\tpicwidthA]{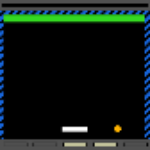} & \includegraphics[width=\tpicwidthB]{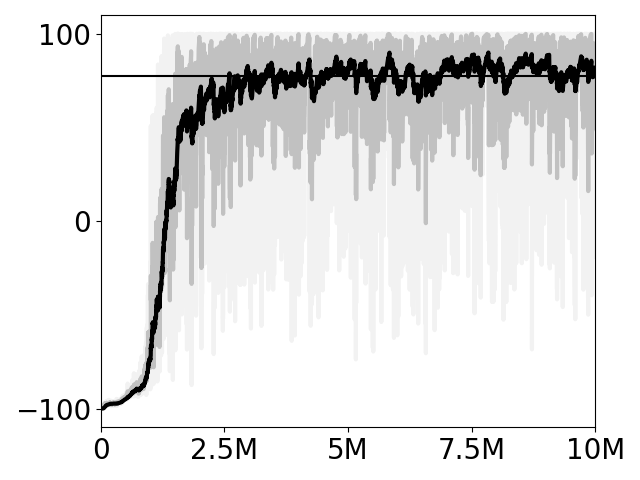} & 77.60 $\pm$ 51.14 \\  
 \hline
 
 Breakout \includegraphics[width=\tpicwidthA]{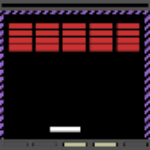} & \includegraphics[width=\tpicwidthB]{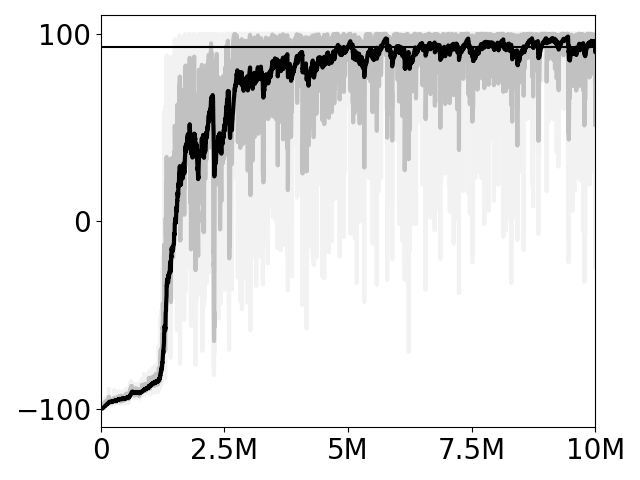} & 93.02 $\pm$ 28.06 & 
 Last Block \includegraphics[width=\tpicwidthA]{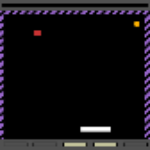} & \includegraphics[width=\tpicwidthB]{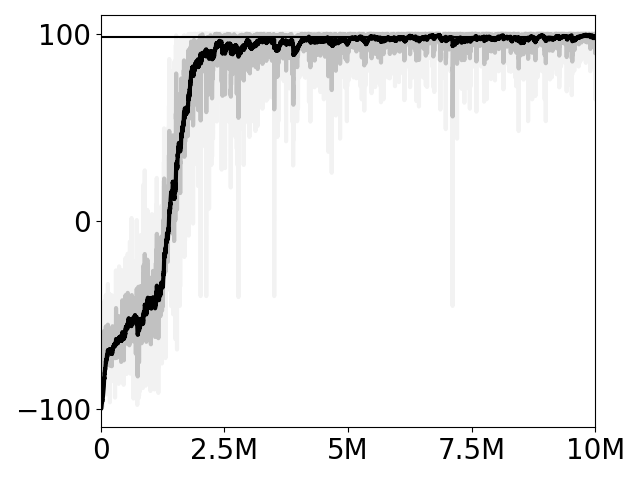} & 98.02 $\pm$ 7.03 \\  
 \hline

 Collect Five \includegraphics[width=\tpicwidthA]{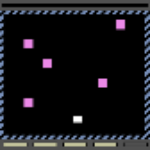} & \includegraphics[width=\tpicwidthB]{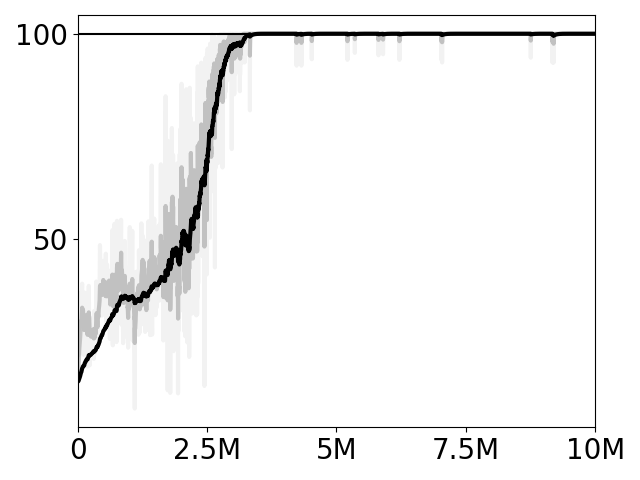} & 100.00 $\pm$ 0.00 & 
 Pong \includegraphics[width=\tpicwidthA]{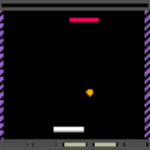} & \includegraphics[width=\tpicwidthB]{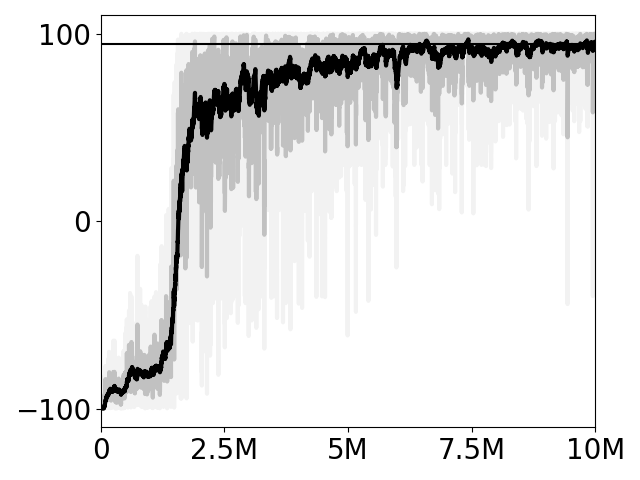} & 94.32 $\pm$ 22.35 \\  
 \hline 
 
 Dodgeball Duel \includegraphics[width=\tpicwidthA]{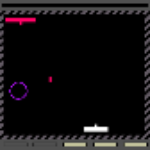} & \includegraphics[width=\tpicwidthB]{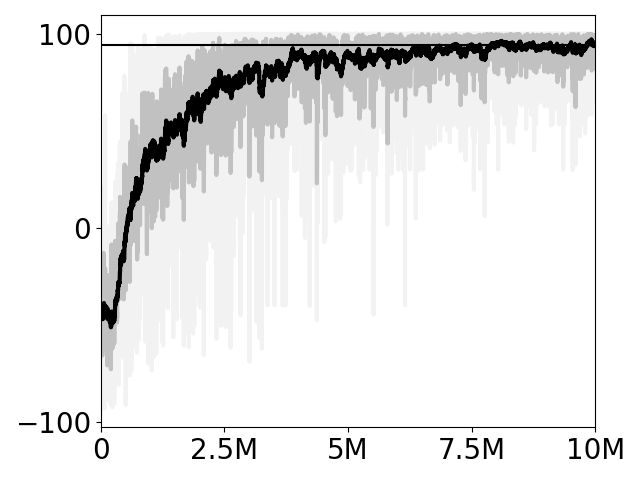} & 94.21 $\pm$ 13.43 & 
 Pong Breakout \includegraphics[width=\tpicwidthA]{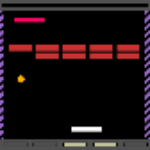} & \includegraphics[width=\tpicwidthB]{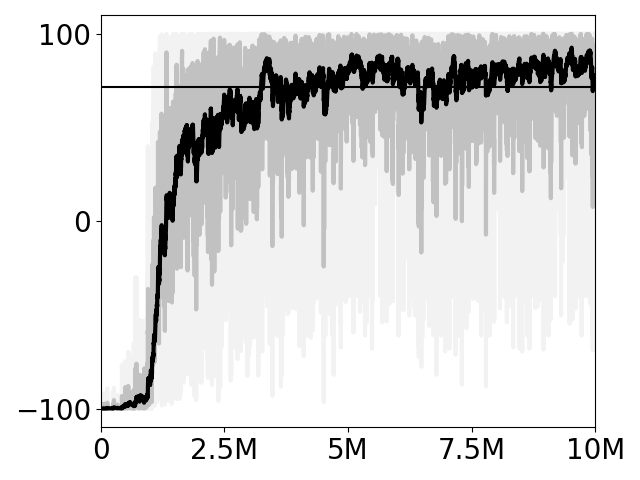} & 71.33 $\pm$ 53.18 \\  
 \hline 

 Duel \includegraphics[width=\tpicwidthA]{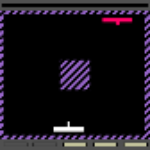} & \includegraphics[width=\tpicwidthB]{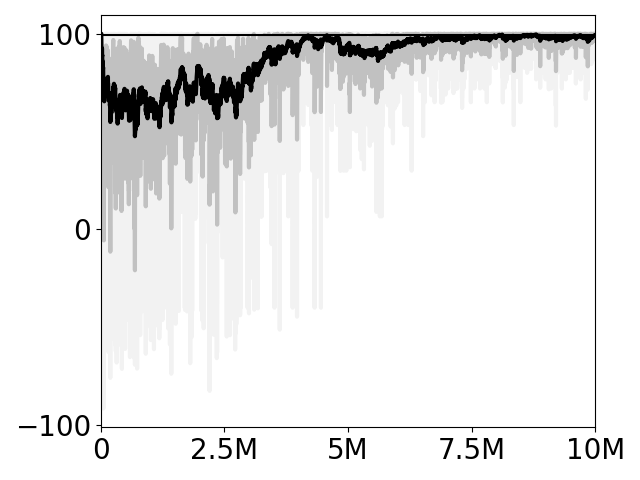} & 99.23 $\pm$ 3.80 & 
 Seek Destroy \includegraphics[width=\tpicwidthA]{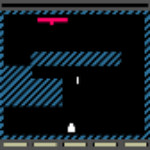} & \includegraphics[width=\tpicwidthB]{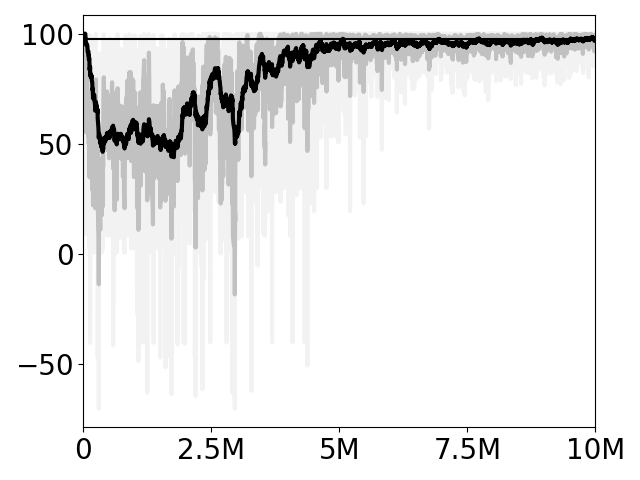} & 97.65 $\pm$ 5.32 \\  
 \hline

 Erosion \includegraphics[width=\tpicwidthA]{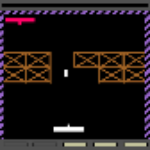} & \includegraphics[width=\tpicwidthB]{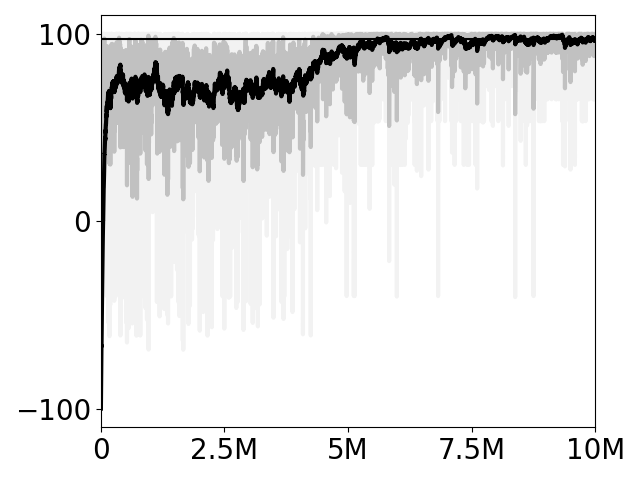} & 97.43 $\pm$ 10.33 & 
 Shootout \includegraphics[width=\tpicwidthA]{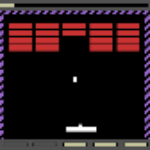} & \includegraphics[width=\tpicwidthB]{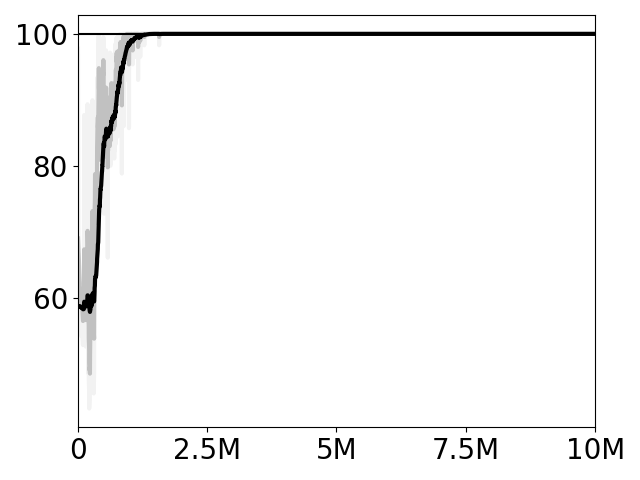} & 100.00 $\pm$ 0.00 \\  
 \hline

 Hedge Maze \includegraphics[width=\tpicwidthA]{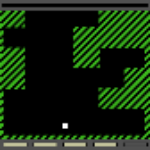} & \includegraphics[width=\tpicwidthB]{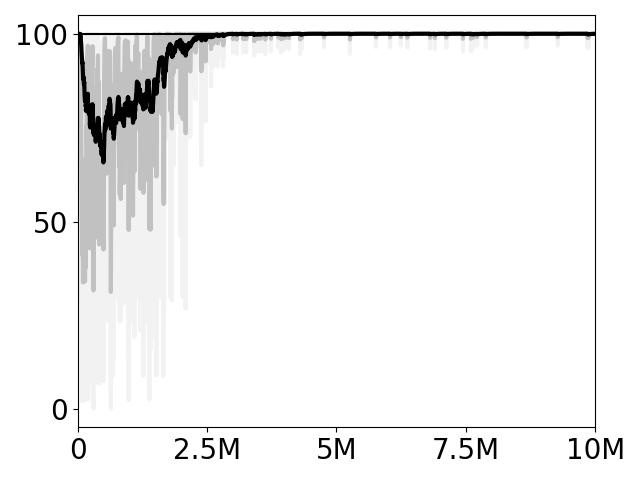} & 100.00 $\pm$ 0.00 & 
 Sweeper \includegraphics[width=\tpicwidthA]{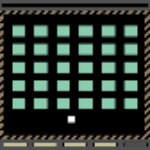} & \includegraphics[width=\tpicwidthB]{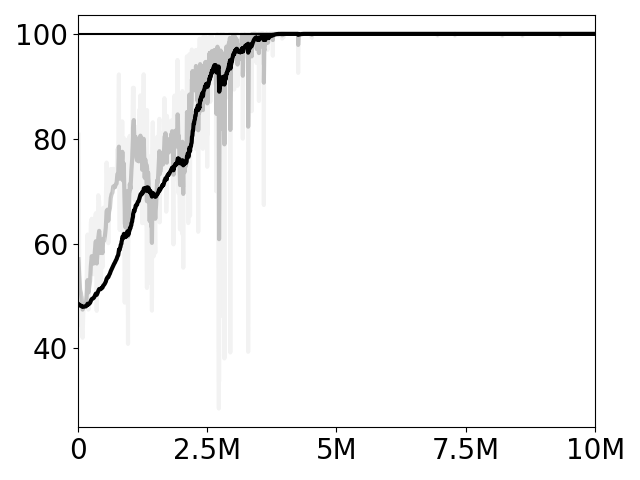} & 100.00 $\pm$ 0.00 \\  
 \hline
 
 Interference \includegraphics[width=\tpicwidthA]{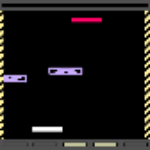} & \includegraphics[width=\tpicwidthB]{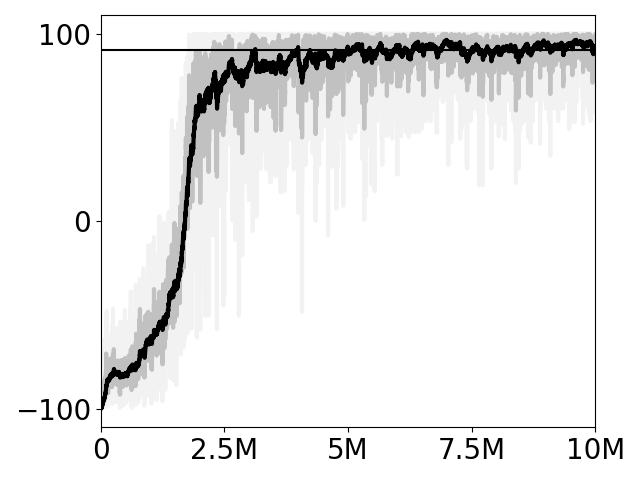} & 91.12 $\pm$ 15.64 & 
 Target Practice \includegraphics[width=\tpicwidthA]{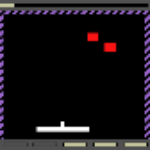} & \includegraphics[width=\tpicwidthB]{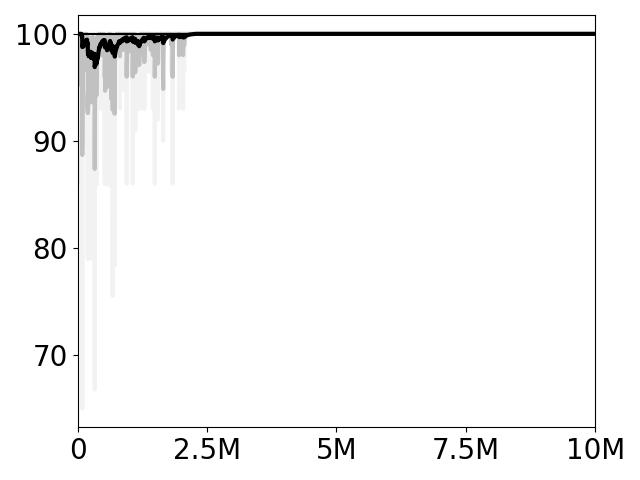} & 100.00 $\pm$ 0.00 \\  
 \hline

 \hline
\end{tabular} \\
\captionsetup{width=0.95\linewidth,skip=0.3cm}
\caption{Predefined games that are clearly solvable with PPO, and their training performance. Different shades of gray show different amounts of smoothing.}
\label{table:1}
\end{table}

\begin{table}[h!]
\centering
\begin{tabular}{|>{\footnotesize\centering\arraybackslash}m{2.4cm}|>{\footnotesize\centering\arraybackslash}m{3cm}|>{\footnotesize\centering\arraybackslash}m{1cm}||>{\footnotesize\centering\arraybackslash}m{2.4cm}|>{\footnotesize\centering\arraybackslash}m{3cm}|>{\footnotesize\centering\arraybackslash}m{1cm}|} 
 \hline
 Game & Training Curve & Score & Game & Training Curve & Score\\
 \hline\hline
 
 Dungeon \includegraphics[width=\tpicwidthA]{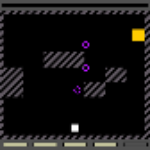} & \includegraphics[width=\tpicwidthB]{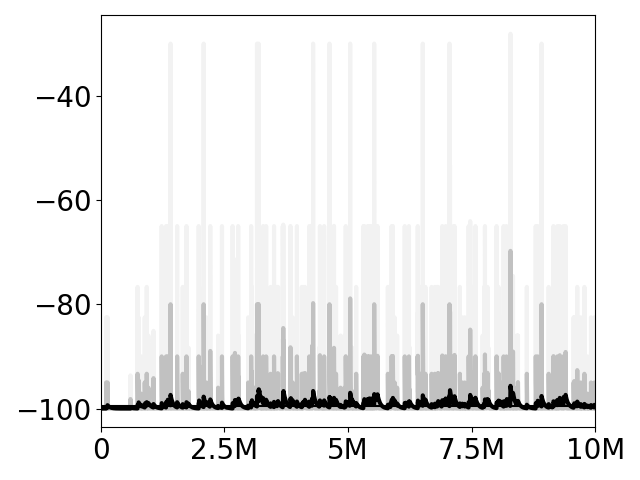} & -99.50 $\pm$ 3.50 & 
 Invasion \includegraphics[width=\tpicwidthA]{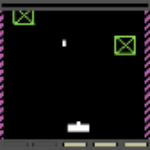} & \includegraphics[width=\tpicwidthB]{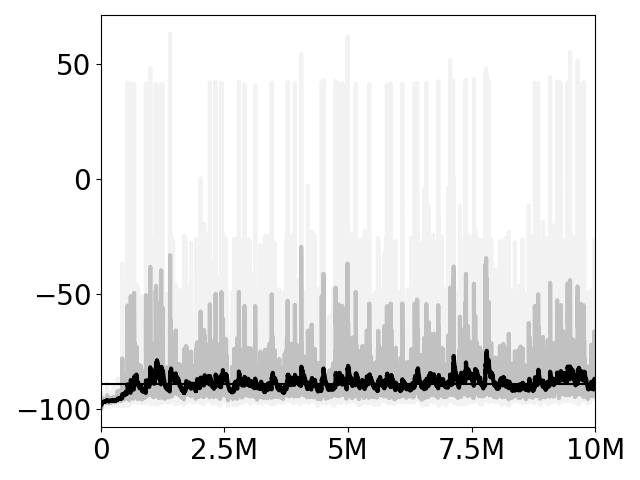} & -89.17 $\pm$ 16.91 \\ 
 \hline

 Freeway \includegraphics[width=\tpicwidthA]{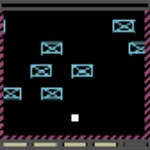} & \includegraphics[width=\tpicwidthB]{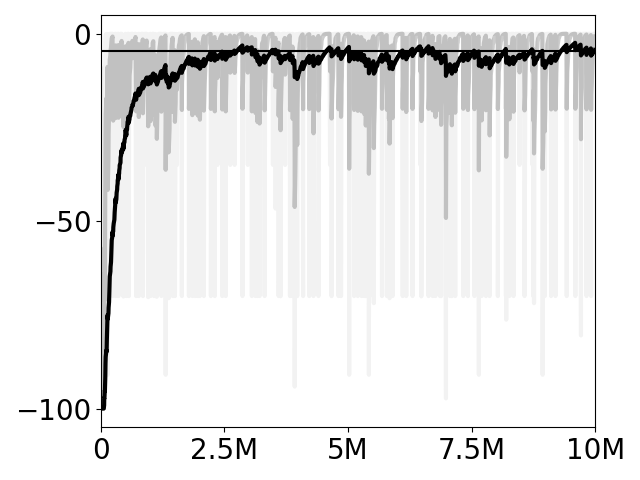} & -4.50 $\pm$ 20.12 & 
 Lava Maze \includegraphics[width=\tpicwidthA]{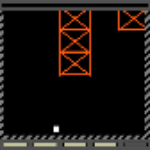} & \includegraphics[width=\tpicwidthB]{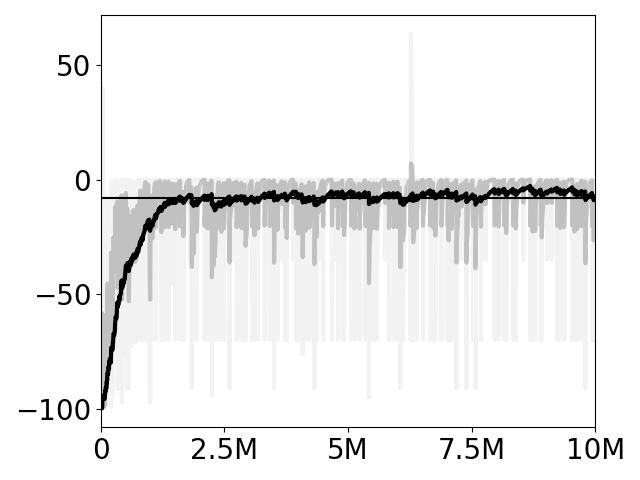} & -8.00 $\pm$ 27.13 \\  
 \hline
 
 Haunted Hallway \includegraphics[width=\tpicwidthA]{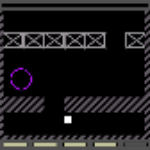} & \includegraphics[width=\tpicwidthB]{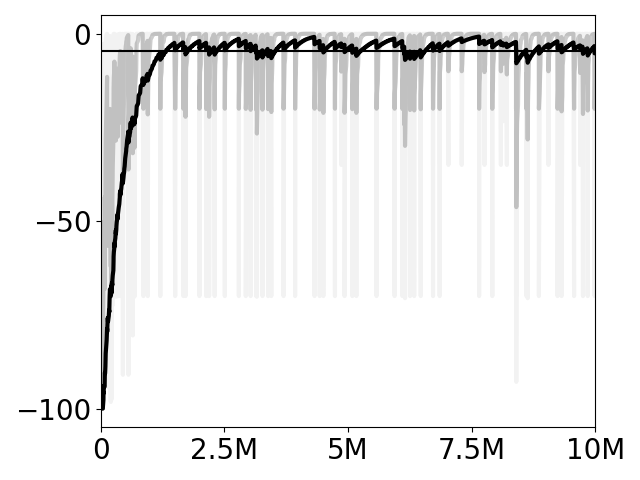} & -4.50 $\pm$ 20.12 &
 Tunneler \includegraphics[width=\tpicwidthA]{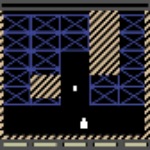} & \includegraphics[width=\tpicwidthB]{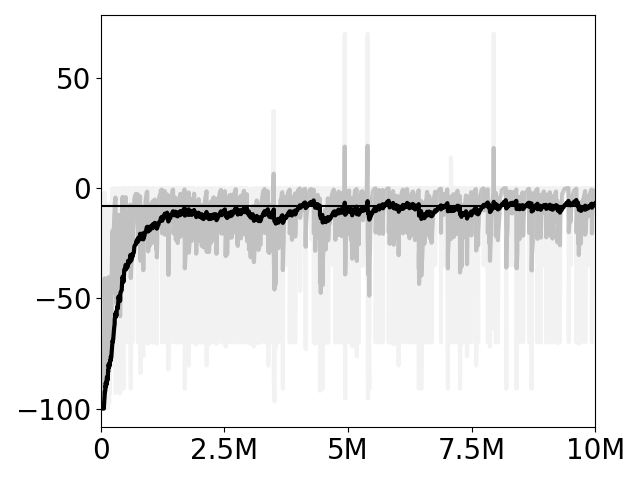} & -8.00 $\pm$ 26.19 \\  
 \hline
\end{tabular} \\
\captionsetup{width=0.95\linewidth,skip=0.3cm}
\caption{Challenge games and their single-task PPO performance.}
\label{table:2}
\end{table}

\subsection{Curricula}

\subsubsection{Curricula Experiments}
In an effort to resolve the learning of challenging problems seen in benchmarking and demonstrate some advanced features of the Meta Arcade environment suite, we conducted a set of experiments that focus on curriculum-based approaches.  Specifically, we designed learning curricula for Dungeon, Freeway, Invasion, Lava Maze, and Tunneler which span the original 10 million training frames but introduce the games’ full complexities over time. These games all fell into the unsuccessful category and merited a curriculum-based approach; other games would probably also benefit from a curriculum, but these had the most room for improvement. Haunted Hallway is omitted because it was unclear how to create a curriculum that did not change the game’s design. The results of the curriculum experiments can be found in Table \ref{table:3}

The general form of each curriculum was to start with what we believe to be an easier set of parameters that would allow early success (in other words, parameters that allow reward discovery with random actions). For example, the Freeway curriculum starts with a distribution of blocks that are small and slow or static, and sometimes has no obstacles at all. Over the first 4 million frames, this parameter distribution is gradually expanded and interpolated to include increasingly difficult parameters during training, but without removing the easier versions from the set of possibilities. This was a conscious decision to allow the algorithm to experience difficult settings while still seeing trivial ones that describe the general goal and reinforce the most primitive idea of the game, such as crossing the screen. From 4 million to 8 million frames, the easy parameters are slowly excluded to focus in on the specific parameters that describe the full game. Some difficulty may still be introduced during this period to completely match the full game. From 8 million steps onward, the original game is played with no alteration.  Some scalar values like speeds and sizes are directly interpolated during the first 8 million frames rather than increasing and then shrinking the sampling range. The last entries in Table \ref{table:3} illustrate how the games may change during training according to the high-level schedule described above.

\subsubsection{Curricula Results}
The curricula approach proved very successful in training most of the challenging games. All the games show final performances that reflect a high level of skill. The Dungeon policy will occasionally fall into positions where the player is irrecoverably surrounded by enemies, and the Invasion policy will sometimes encounter a pattern of falling blocks that requires perfect movement back and forth to either edge of the screen.  However, in the large majority of games these policies are very successful, with average scores far above their PPO benchmark counterparts.

The key trend during training seems to be that the policy quickly reaches strong play on the easy games, and then tries to maintain high performance as the game parameters become more difficult. The Invasion curriculum leads to a slow ramp-up in score since the early games have fewer blocks than are needed to end the game. A score of 100 is not actually possible until further into the curriculum. Several games also show a performance dip that coincides with a switch to the final game, which may indicate that the transition to the full game should have been drawn out over more frames for stability. For example, Tunneler takes a sharp dive at 8M frames but is able to recover. If the curriculum were more compressed, it is possible that recovery would not have occurred.

The curricula experiments examine changes within a single task, but it is possible that the needed skills could be acquired from completely separate games. By design, Meta Arcade environments have a great deal of overlap. The final experiments, discussed in the next section, take advantage of this property to explore transfer between selected games.

\newcommand{\tpicwidthC}{8cm}

\begin{table}[h!]
\centering
\begin{tabular}{|>{\footnotesize\centering\arraybackslash}m{2.4cm}|>{\footnotesize\centering\arraybackslash}m{8cm}|>{\footnotesize\centering\arraybackslash}m{1cm}|} 
 \hline
 Game & Curriculum Training Curve & Final Score\\
 \hline\hline
 
 Dungeon \includegraphics[width=\tpicwidthA]{game_stills/dungeon.png} & \includegraphics[width=\tpicwidthC]{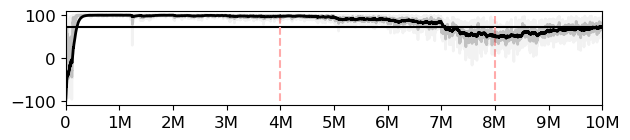} & 72.08 $\pm$ 17.51 \\ 
 \hline

 Freeway \includegraphics[width=\tpicwidthA]{game_stills/freeway.png} & \includegraphics[width=\tpicwidthC]{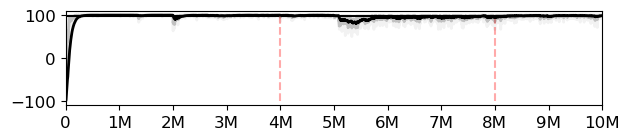} & 98.67 $\pm$ 4.88 \\  
 \hline
 
 Invasion \includegraphics[width=\tpicwidthA]{game_stills/invasion.png} & \includegraphics[width=\tpicwidthC]{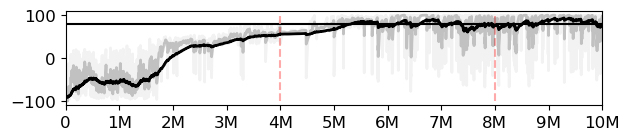} & 79.56 $\pm$ 53.86 \\  
 \hline
 
 Lava Maze \includegraphics[width=\tpicwidthA]{game_stills/lava_maze.png} & \includegraphics[width=\tpicwidthC]{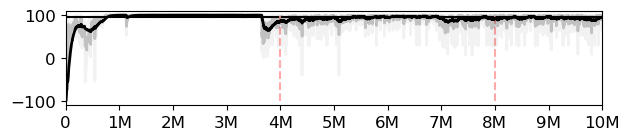} & 95.52 $\pm$ 13.55 \\  
 \hline
 
 Tunneler \includegraphics[width=\tpicwidthA]{game_stills/tunneler.png} & \includegraphics[width=\tpicwidthC]{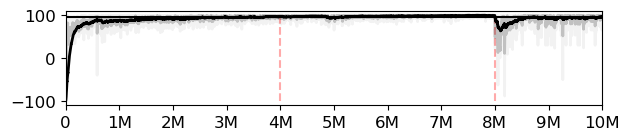} & 95.95 $\pm$ 10.26 \\  
 \hline
 
Tunneler Example Frames & \includegraphics[width=8cm]{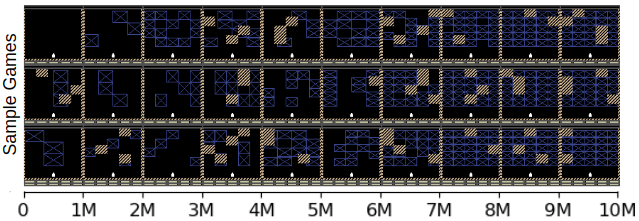} & - \\ 
 \hline

Curriculum Sampling Strategy & \includegraphics[width=7.6cm]{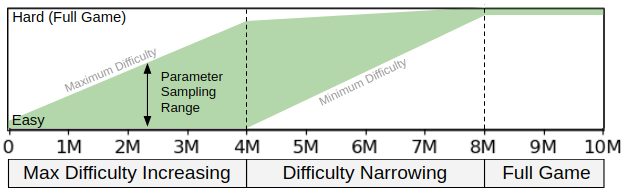} & - \\
 \hline
 
\end{tabular} \\

\captionsetup{width=0.95\linewidth,skip=0.3cm}
\caption{Successful curricula training on several games in which single-task PPO failed. \textbf{Second from bottom:} Example frames sampled from the Tunneler curriculum, showing variety on the y-axis and an increase in difficulty along the x-axis. \textbf{Bottom:} The general approach to curriculum parameter sampling, with a widening of the sampling envelope for 4M frames followed by a shrinking of the envelope to focus on the full game.}
\label{table:3}
\end{table}

\subsection{Multiple Games}

Our final set of experiments serve as exploratory examples of transfer learning between games in Meta Arcade. A major design goal of Meta Arcade was to provide a suite of games with common features and mechanisms, an important precursor to transfer learning via multitask learning or meta-learning. Here we focus on predefined games with minimal alterations, the only exception being that games were given a common color scheme to ease transfer. Future work could examine transfer between color schemes or domain randomization over colors -- both are easily implemented given the configurability of the environment.

A full table of the characteristics of predefined games can be found in the appendix (Table~\ref{table:4}). Each of these games could also be varied in appearance and dynamics, and it does not necessarily include characteristics of potential new games defined to study transfer.

\subsubsection{Transfer Experiments}
We first select three predefined games which we believe require learning skills that could readily transfer to other games.  The three games are learned together with a single PPO policy (multitask learning) for 50 million steps. We use a model checkpoint from 30 million steps as a starting point for additional games that may have common skills or features. The decision to use the 30 million step checkpoint was inspired by \cite{actormimic}, which notes that training for a long time may lead to overfit policies that that do not transfer well between tasks. Unlike \cite{actormimic}, we did not re-initialize the final layer or change the model size, as this experiment is primarily exploratory. A future study could examine the optimal point during training to transfer weights, which weights to transfer, and how large the model should be. When attempting transfer, the new games were trained for 1 million steps, as in \cite{sonic}.

\begin{figure}[ht!]
\centering
  \includegraphics[width=\linewidth]{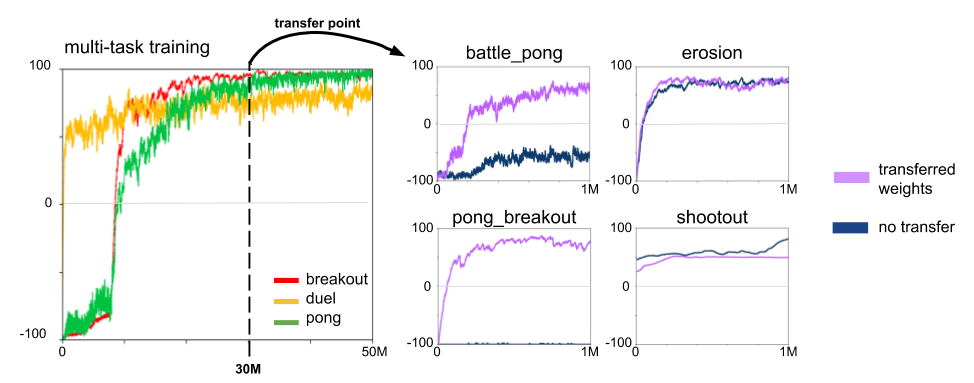}
  \captionsetup{width=0.95\linewidth}
  \caption{\small Multitask learning and transfer among selected paddle-based games.}
  \label{fig:exp31}
\end{figure}

The first experiment examined transfer between paddle-based games. The three multitask games were Breakout, Duel, and Pong, which demonstrate a variety of game objectives that may include a paddle for the player. These three tasks were trained jointly to create a single policy that performed well on all three games. Transfer from this policy was attempted to Battle Pong, Erosion, Pong Breakout, and Shootout independently. These results can be seen in Figure \ref{fig:exp31}.

The second experiment examined recombination of more diverse skills, including two-dimensional movement. The three multitask games were Collect Five (demonstrating block collection), Hedge Maze (demonstrating the traversal objective), and Erosion (demonstrating shooting). Transfer was attempted to Avalanche, Dungeon (without hazards), Seek Destroy, and Tunneler. These results can be seen in Figure \ref{fig:exp32}. A full list of objectives is found in the appendix, Table \ref{table:4}.

\begin{figure}[ht!]
\centering
  \includegraphics[width=\linewidth]{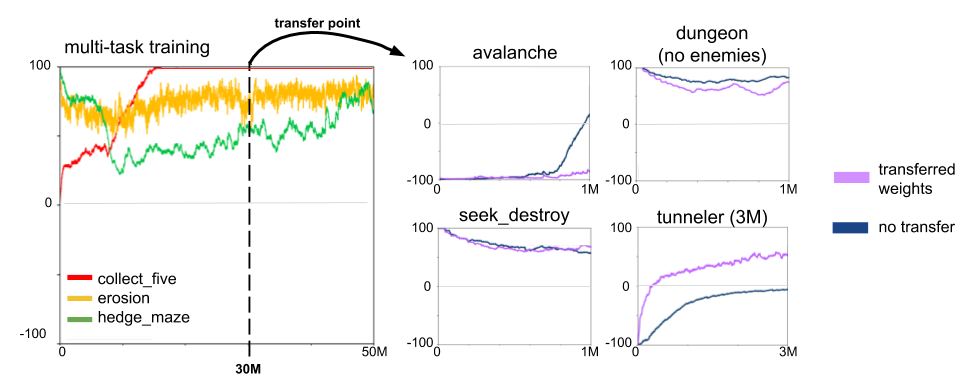}
  \captionsetup{width=0.95\linewidth}
  \caption{\small Multitask learning and transfer among selected games demonstrating a range of skills.}
  \label{fig:exp32}
\end{figure}

\subsubsection{Transfer Results}

The paddle-based experiments showed clear transfer among the games using a bouncing ball. This mechanism appears in both Breakout and Pong, which were closely tied during multitask training, indicating that this common skill was possibly learned in a generic way to satisfy both games. When transferring to Battle Pong and Pong Breakout, there was a significant benefit over starting from scratch. There was little benefit when transferring to Erosion or Shootout, which are shooting-based games. However, these games are learned incredibly quickly and perhaps a more difficult game was needed to visualize the importance of transfer.

The second experiment using a more diverse range of skills had less explainable results. Several of the training curves exhibit a drop in performance before potentially improving; this was also seen in some of the initial PPO benchmarking. We believe this is probably due to high success with a random policy that initially loses performance as it converges towards a learned policy with less entropy, but this needs further investigation. There was no observed benefit to transfer aside from the game Tunneler, which was trained for 3 million steps instead of 1 million to flush out this result. This was a surprising result, as Tunneler is one of the more challenging predefined games in the suite, appearing unsolvable by PPO in baselining. Presumably, the combination of Hedge Maze (reach the other side) and Erosion (destroy blocks) allowed the transfer learner to more readily experience the goal, which lead to successful training. This indicates that if basis games are chosen correctly and properly transferred, there is potential to learn some very powerful policies. How to best choose what games to include in a meta-learning curriculum is an area of research worthy of further exploration with Meta Arcade.

\section{Conclusions and Future Work} \label{future}
In this paper we presented Meta Arcade, a tool to create 2D arcade games that are built from common mechanisms, such that skills or learned features from one game may be applicable to others. The games are highly customizable by changing a common set of parameters, enabling a vast array of variations for any one game or even games defined by distributions. We provide a set of single-task learner benchmarks on the games using our own implementation of PPO, and then show how taking advantage of Meta Arcade's features and curriculum learning enable successful learning on tasks where single-task learning failed. Finally, we summarized initial results demonstrating successful policy transfer from sets of Meta Arcade games to other games in the suite. Meta Arcade enables research into curriculum learning, multitask learning and meta-learning for RL, as partially illustrated in some of our experiments. 

Future work could use Meta Arcade to characterize transfer between tasks, explore when transfer is possible, or look to continual learning for sequences of tasks. Our curriculum-based experiments showed a clear benefit to curricula that lead a policy through environment configurations, and future work formalizing this into pedagogy would be beneficial for problems where such configurability is possible. The final experiment initialized the learning of Tunneler with a policy that could learn successfully -- future work could examine if this is possible for other games, and determine when such initialization is possible through transfer learning. In the meta-learning domain, Meta Arcade provides a set of overlapping tasks that use discrete actions and have sparse rewards, an area which has proven difficult for existing meta RL algorithms. It is unclear to what extent could game parameters be changed and be viable for adaptation under techniques like MAML \cite{maml} and Reptile \cite{reptile}.  The environment suite itself may evolve over time, with features added as needed to support the latest promising areas in DRL research.

\section*{Disclaimer}
Any opinions, findings, and conclusions or recommendations expressed in this material are those of the author(s) and do not necessarily reflect the views of the Office of Naval Research. The views, opinions, and/or findings expressed are those of the author(s) and should not be interpreted as representing the official views or policies of the Department of Defense or the U.S. Government.

\clearpage

\bibliographystyle{unsrt}  
\bibliography{references}  

\pagebreak
\appendix

\section{Appendix}
\subsection{PPO Implementation and Hyperparameters} \label{ppoparams}

In the benchmarking and curriculum experiments, the following hyperparameters were used for PPO:

\begin{center}
\begin{tabular}{ |c|c| } 
    \hline
    Parameter & Value \\
    \hline
    \hline
    Trajectory Length & 128 \\ 
    \hline
    Parallel Trajectories & 8 \\
    \hline
    Epochs & 3 \\
    \hline
    Clipping Parameter & 0.1 \\
    \hline
    Adam Learning Rate & 0.00025 \\
    \hline
    Entropy Coefficient & 0.01 \\
    \hline
    Value Coefficient & 1.0 \\
    \hline
    Maximum Gradient Norm & 0.5 \\
    \hline
    Gamma & 0.99 \\
    \hline
    Lambda & 0.95 \\
    \hline
\end{tabular}
\end{center}
\captionsetup{width=0.95\linewidth,skip=0.3cm}
\captionof{table}{PPO Hyperparameters}
\label{table:hyperparams}
\bigskip

Parameters such as the clipping parameter and entropy coefficient were held fixed during training, and not annealed towards zero as in some implementations.

\subsection{Task Boundary Discussion}
When the parameters of a task are randomized or modifiable, as in Meta Arcade, there is a gray area between varying the parameters of the game and introducing a separate task.  The configurablility of the environment suite makes it easier to investigate this boundary. In our experiments, this was especially relevant to curriculum learning: while the task slowly changed over time, the start and end configurations often looked quite different from each other. Most curricula for games with the objective to cross the screen had a common starting point -- the possibility of a blank play area with no obstacles. If this were defined as a separate task then these curricula could be considered multi-task problems with at least one task in common. Similarly, the curriculum for Invasion started with blocks that were collectable instead of harmful, in order to encourage positioning the paddle beneath the falling blocks. This mechanism is also found in the predefined game Avalanche. In general, it is useful to think of the predefined games as data points in the space of possible tasks, and it is up to future researchers to define metrics for similarity and uniqueness.

\subsection{Predefined Games Characteristics}
For reference, we provide a list of predefined games and which environment components they include (Table \ref{table:4}). This is especially useful for determining if any two (or more) games may be candidates for transfer learning.

\begin{landscape}
\begin{table}[ht!]
\centering
\includegraphics[width=22cm]{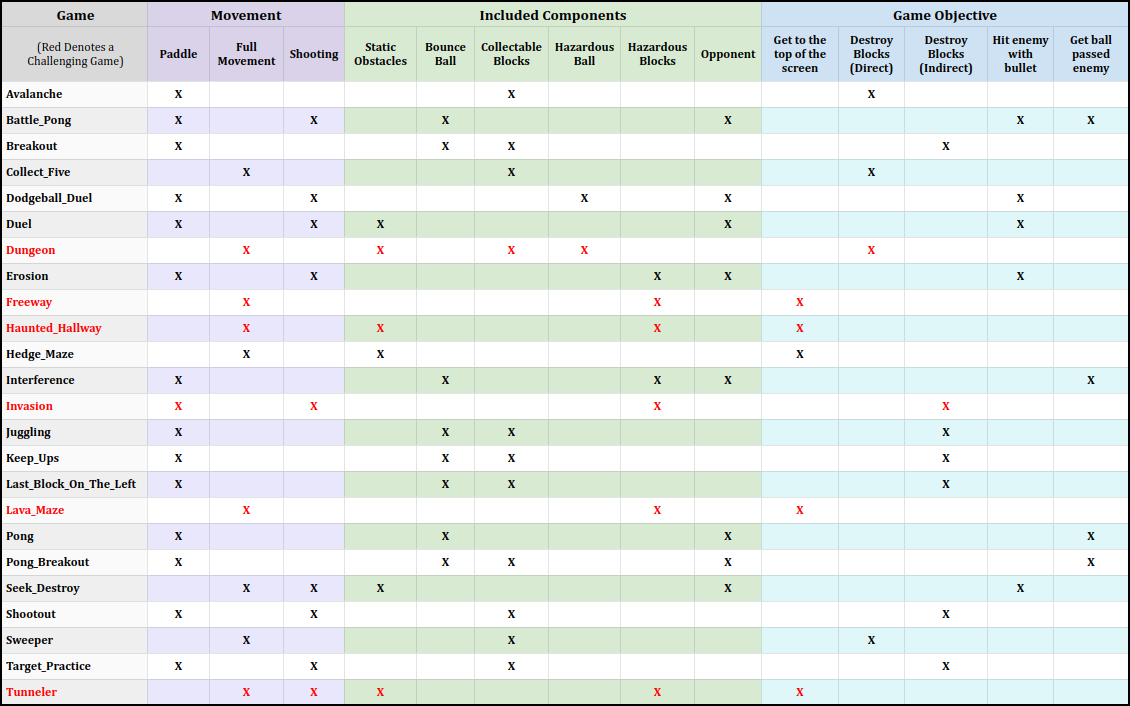}

\captionsetup{width=0.9\linewidth,skip=0.5cm}
\caption{Summary of skills and elements common to the set of predefined games.}

\label{table:4}
\end{table}
\end{landscape}

\subsection{Example Game JSON Definition}

\begin{figure}[ht!]
    \centering
    \begin{lstlisting}[language=Python, basicstyle=\footnotesize]
    {
    	"meta":{
    		"description":"Catch 50 falling blocks in a row."
    	},
    	"actions":{
    		"up":false,
    		"down":false,
    		"left":true,
    		"right":true,
    		"fire":false
    	},
    	"game_elements":{
    		"top_wall":false,
    		"bottom_wall":false,
    		"ball":false,
    		"opponent":false,
    		"blocks":true,
    		"static_barriers":false
    	},
    	"display_settings":{
    		"background_color":[0,0,0],
    		"ui_color":[80,80,80],
    		"indicator_color_1":[200,200,160],
    		"indicator_color_2":[0,0,0]
    	},
    	"player_settings":{
    		"width":0.15,
    		"height":0.05,
    		"speed":0.012,
    		"color":[255,255,255],
    		"steering":0.5
    	},
    	"opponent_settings":{},
    	"ball_settings":{},
    	"blocks_settings":{
    		"creation_area":[0.05,-1.0,0.9,1.0],
    		"rows":6,
    		"cols":6,
    		"per_row":1,
    		"spacing":0.4,
    		"color":[162, 219, 252],
    		"static_weave_fall":"fall",
    		"speed":0.003,
    		"harmful":false,
    		"points":2
    	},
    	"static_barrier_settings":{
    		"color":[38, 101, 209]
    	},
    	"image_settings":{
    		"color_inversion":false,
    		"rotation":0,
    		"hue_shift":0.0,
    		"saturation_shift":0.0,
    		"value_shift":0.0
    	}
    }
    \end{lstlisting}
    \caption{Example JSON of the game \textit{Avalanche}.}
    \label{fig:json_game}
\end{figure}

\end{document}